\documentclass[sn-mathphys,Numbered]{sn-jnl}

\usepackage{graphicx}%
\usepackage{multirow}%
\usepackage{amsmath,amssymb,amsfonts}%
\usepackage{amsthm}%
\usepackage{mathrsfs}%
\usepackage[title]{appendix}%
\usepackage{xcolor}%
\usepackage{textcomp}%
\usepackage{manyfoot}%
\usepackage{booktabs}%
\usepackage{algorithm}%
\usepackage{algorithmicx}%
\usepackage{algpseudocode}%
\usepackage{listings}%
\usepackage{amsthm}
\usepackage{amsmath} 
\usepackage{amsfonts}
\usepackage{amssymb}
\usepackage{booktabs}

\def\bP{\mathbb{P}}

\theoremstyle{thmstyleone}%
\newtheorem{theorem}{Theorem}
\theoremstyle{thmstyletwo}%

\theoremstyle{thmstylethree}%

\usepackage{bbm} 
\usepackage[shortlabels]{enumitem}
\begin{document}

\title[Entropic learning perspective on AI resilience, adversarial attack algorithms and and self-attacking learning.]{Entropic perspective on AI resilience, data-minimal learning and optimal control.}


\author*[1]{\fnm{Illia} \sur{Horenko}}\email{horenko@rptu.de}


\affil*[1]{\orgdiv{Chair for Mathematics of AI, Faculty of Mathematics}, \orgname{RPTU Kaiserslautern-Landau}, \orgaddress{\street{Gottlieb-Daimler-Str. 48}, \city{Kaiserslautern}, \postcode{67663}, \country{Germany}}}

\abstract{
Behind the general terms like \emph{adversarial attacks} and \emph{generative adversarial networks (GANs)},  such criteria as AI resilience and AI robustness gained a crucial importance, in trying to make the emergent AI tools reliable, energy-efficient and safe. From mathematical  perspective, these problems boil down to finding  a mathematically-sound quantification of  the smallest-sufficient amounts of the perturbations of function arguments that lead to the largest possible perturbations of the approximated function values.   
Here, the entropy-optimising perspective on adversarial algorithms from Artificial Intelligence (AI) is proposed to attack this problem.  It is shown that adopting this perspective helps proving  computational conditions for the global optimality and uniqueness of adversarial attacks based on cheaply-verifiable  mathematical criteria. Further, it is shown how this perspective can be used for developing self-attacking learning algorithms, that  generate optimal new data points for training, replacing one of  the trained agents with this mathematical criterium. On a  broad selection of various synthetic and real-life problems from hydrodynamics and biomedicine, it is  shown that such self-attacking learning algorithms allow training orders-of-magnitude simpler and cheaper models with superior performance - and can be used to directly train the optimal controls in complex systems, requiring only a small fraction of the training data and outperforming the as-comprehensive-as-the-author-was-able-to-find set of contemporary AI tools (like boosted Random Forests, deep neural networks, foundational models based on transformer architectures, etc.) in complexity (measured as the model descriptor length) and predictive performance.} 
\maketitle

\section{Introduction}
\label{sec:introduction}

Understanding and measuring accountability and robustness of rapidly-developing Artificial Intelligence tools (AI) is recently moving to the center of AI research. As formulated in the seminal paper featuring the sparks of artificial general intelligence that are shown by Chat GPT-4, in the last sentence of their paper the Microsoft Research team says that "elucidating the nature and mechanisms of AI systems such as GPT-4 is a formidable challenge that has suddenly become important and urgent" \cite{sparks23}. Importance of getting a better understanding of the nature and mechanisms of the AI is especially underlined by the multiple very spectacular recent examples  of the so-called adversarial attacks on AI tools, when very small changes in the input data could be used for a complete corruption of the AI classification outcomes \cite{adverse2}. This idea of creating a pair of learning agents - the one that tries to learn the problem and another that creates the new learning data points that are most challenging for the first agent - is in the core of the rapidly-developing field of  Generative Adversarial Neuronal Networks (GANNs) and reinforcement learning algorithms for optimal control problems \cite{liu03,ziebart2008maximum,manandhar14,ho2016generative,pfau2016connecting,haarnoja2018soft,gu19,li21b,nguyen21,aggarwal21,adverse2,li21,adverse1,wang23}. Similar adversarial attacks were recently shown to help designing strategies that beat the superhuman AlphaGo AIs - showcasing significant problems with robustness of such AI tools  \cite{adverse3}. On the mathematical side of this research, it was demonstrated that one can use the same mathematical concepts, like the gradient descent method (normally used in the training of AI tools), to find the adversarial attacks as solutions of numerical optimisation problems \cite{adverse1}.

Despite of many reported success-stories, the main mathematical and numerical challenges in this area remain quite significant, including: (i) mathematical restrictions like the Lipschitz-continuity of the AI decision function derivatives (required for the convergence of the AI training algorithms, see, e.g., eq. 4 and Theorem 1 in \cite{adverse1}); (ii) practical impossibility to verify if the adversarial attacks obtained with such algorithms are globally-optimal and can not be improved; (iii)  slow convergence, high cost and data-greediness, requiring extensive computational resources and large amounts of training data, and (iv) necessity of simultaneously training two competing learning agents instead of one that is required in the more classical AI tools. 

The main message of this manuscript is in showing that deploying the entropic perspective on the problem, e.g., as adopted by the entropic AI methods \cite{Horenko_2020,horenko_pnas_22,espa_22,horenko_pnas_23}, allows a robust and polynomially-scalable  computation of such adversarial attacks, as well as a direct learning in optimal control problems. The role of the second learning agent is then taken  by the procedure based on a direct application of mathematical conditions for existence and uniqueness of obtained solutions based on straightforward computational criteria. This strategy requires only one learning agent, since the second one (generating new data points that aim to challenge the first one) is replaced by this computational criterium, that generates the optimal new training data points with the maximal entropy in terms of the learning agent.

\section{Methods}
\label{sec:methods}
\subsection{Mathematical formulation of the adversarial attacks on learning tools}
\label{sec:m_formulation}
From mathematical perspective, AI classification tools can be considered as functions, that for a given argument $X$ (called a feature vector) assign a probability vector  $(\bP_{1}(X)),\dots,\bP_{M}(X))$  of probabilities to belong to one of the  $M$ label classes. Training such AI classifier on training data pairs $(X,L)$ (where $L\in\mathbb{L}$ denotes a known label of the given data instance $X$) means approximating the mapping $X\rightarrow \mathbb{L}$ (where $\mathbb{L}$ is a set of labels), on the set of the available data with the function  $\bP_L(X)$ (called the affiliation probability function). Different AI learning tools result in the different approximations of the underlying mapping $X\rightarrow \mathbb{L}$, and in the different affiliation probability functions.  
In the so-called untargeted attacks, for a given feature vector $X$ with a label $L$ and an AI classifier with the trained class affiliation probability function $\bP_L(X)$, one is trying to find the adversarial vector $X^*$ that is as close as possible to the given vector $X$  in terms of some distance measure $\text{dist}(X,X^*)$ and such that $\bP_{L^*}(X^*)$ is attaining its maximal value for $L^*\neq L$ \cite{adverse2,adverse1}. 
In many AI applications, only a subset $d_a\subseteq\left\{1,\dots,D\right\}$ 
of the $D$ degrees of freedom of the feature vector $X$ is available for changes and control in the adversarial attack - and there is a complementary feature subset $d_{na}=\left\{1,\dots,D\right\}/d_a$ (e.g., an age or a sex of the patient) that are not accessible for such an attack or control.   For a given  feature vector $X$ with label $L$, accessible features subset $d_a$, and a class affiliation probability function $\bP_L(X)$,  in the following we will consider finding minimal adversarial perturbation (MAP) $(X,X^*)$, connecting $X$ to (a priori unknown) $X^*$ from a solution of the following optimisation problem: 
\begin{eqnarray}
\label{eq:L}
X^*&=&\arg\min_{\tilde{X}}\text{dist}(X,\tilde{X}),\\
\label{eq:L_con}
\text{such that }&&\bP_{L}(X)-\bP_{L}(\tilde{X})\geq\delta>0.\\
\label{eq:L_con_ind}
\tilde{X}_{i\in d_{na}}&=&{X}_{i\in d_{na}}.
\end{eqnarray}
Minimal adversarial distance measure $MAD_{L}(X,\delta)=\text{dist}(X,X^*)$ will then be defined as a size of this perturbation.

\subsection{Entropic learning with the entropy-optimal Sparse Probabilistic Approximation algorithm (eSPA+)}
\label{sec:espa}

 Entropy-optimal Scalable Probabilistic Approximation  \cite{Horenko_2020,Gerber_2020,espa_22,horenko_pnas_22,horenko_pnas_23,horenko_24a,horenko_24b} is formulated and implemented as a minimisation of the following learning functional $\mathcal{L}_{\textrm{eSPA+}}$ for the real-valued training data feature matrix $\tilde{X}\in\mathcal{R}^{D\times T}$ (where $T$ is the size of the training statistics) and training stochastic matrix of label probabilities $\tilde{\Pi}\in\mathcal{P}^{M\times T}$ (where $M$ is the number of label classes):    
\begin{eqnarray}
\label{eq:L_tilde}
\mathcal{L}_{\textrm{\textrm{eSPA+}}}&=&\underbrace{\frac{1}{T}\sum_{d=1}^DW_d\sum_{t=1}^T\sum_{k=1}^K\Gamma_{kt}(\tilde{X}_{dt}-S_{dk})^2}_{\textrm{Jensen's bound on feature discretization}} +\epsilon_E\underbrace{\sum_{d=1}^DW_d\log(W_d)}_{\textrm{entropic feature sparsification}}\nonumber\\
&-&\frac{\epsilon_{CL}}{T}\underbrace{\sum_{m,t=1}^{M,T}\tilde{\Pi}_{mt}\sum_{k=1}^K\Gamma_{kt}\log\left(\Lambda_{mk}\right)}_{\textrm{Jensen's bound on KL-divergence between true and trained labels}},\nonumber\\
\left[ W^*,\Gamma^*, \Lambda^*, S^* \right]&=&\arg\min_{W,\Gamma, \Lambda, S}\mathcal{L}_{\textrm{eSPA+}}\nonumber\\
\text{such that } W\in\mathcal{P}^{D\times 1}, \Gamma\in\mathcal{P}^{K\times T}&,& \Lambda\in\mathcal{P}^{M\times K} \text{ are stochastic matrices, and } S\in\mathcal{R}^{D\times K}.
\end{eqnarray}
Here, $W$ is the vector of feature dimension weights, $S$ is the matrix with columns being Voronoi cell centres, $\Gamma$ is the stochastic matrix of probabilities for the training data to occupy certain Voronoi cells, and $\Lambda$ is the matrix of conditional probabilities connecting label probabilities and probabilities to occupy certain Voronoi cells.  The first term of $\mathcal{L}_{\textrm{eSPA+}}$ in (\ref{eq:L_tilde})  is the upper bound on the convex approximation $\tilde{X}_{d,t}\approx\sum_{k=1}^K\Gamma_{kt}S_{d,k}$ of features when applying the Jensen's inequality. The third term represents Jensen’s upper bound for the Kullback-Leibler divergence between the true label probabilities $\tilde{\Pi}_{mt}$ and their approximations $\sum_{k=1}^K\Lambda_{mk}\Gamma_{kt}$ that are representing the exact law of total probability:  it relates probability measures $\tilde{\Pi}$ for observed labels and probabilities $\Gamma$ (for features to occupy Voronoi cells) by means of the rectangular stochastic matrix $\Lambda$ of conditional probabilities \cite{horenko17}. 

 As proven in  \cite{espa_22}, for given training data $\left[\tilde{X},\tilde{\Pi}\right]$, and fixed values of three scalar hyper-parameters $\left[K,\epsilon_E,\epsilon_{CL}\right]$, eSPA model parameters $\left[ W^*,\Gamma^*, \Lambda^*, S^* \right]$ in (\ref{eq:L_tilde}) can be trained with the iterative computational cost scaling $\mathcal{O}(DKT)$ - i.e., with the same cost scaling as of the very popular K-means clustering algorithm.
\subsection{Mathematics of the adversarial attacks on eSPA+}
\label{sec:espa_attack}
  As shown in   \cite{Horenko_2020}, for any (yet-unlabelled) feature vector $X\in\mathcal{R}^{D\times 1}$ and any $L\in\left\{1,\dots,M\right\}$,  eSPA+ classifier decision function  is a piecewise-constant function confined by the boundaries of the  $K$ Voronoi cells (cell centres  are given by the columns of the matrix $S^*$):
 \begin{eqnarray}
\label{eq:bP_espa}
 \bP^{\textrm{\textrm{eSPA+}}}_L(X)&=&\Lambda^{*}_{L,k^*}, \textrm{ where } k^* = \arg\min_{k}\sum_{d=1}^DW^*_d\sum_{t=1}^T(X_{d}-\left\{{S^*}\right\}_{dk})^2.
 \end{eqnarray}
 As can be seen from (\ref{eq:bP_espa}), inside each Voronoi cell $k$  in the $W$-weighted Euclidean space, the value $\bP^{\textrm{eSPA}}_L(X)$ remains constant and equal $\Lambda^{*}_{L,k}$ in the whole interior of this cell - and the cell boundaries between any pair of  cells $k$ and $k'$ (where the value can change from $\Lambda^{*}_{L,k}$ to $\Lambda^{*}_{L,k'}$) are defined by a hyperplane orthogonal to the following normalized vector $V^{k,k'}$, connecting the cell centres $S^{*}_{:,k}$ and  $S^{*}_{:,k'}$: 
 \begin{eqnarray}
\label{eq:Voronoi_vec}
V^{k,k'}&=&\frac{\left(\left(W^*\right)^{0.5}\circ S_{:,k}-\left(W^*\right)^{0.5}\circ S_{:,k'}\right)}{\left\|\left(W^*\right)^{0.5}\circ S_{:,k}-\left(W^*\right)^{0.5}\circ S_{:,k'}\right\|_2},
 \end{eqnarray}
and going through a midpoint  
\begin{eqnarray}
\label{eq:Voronoi_mid}
V_{mid}^{k,k'}&=&0.5\left(W^*\right)^{0.5}\circ\left(S_{:,k}+S_{:,k'}\right).
 \end{eqnarray}
 Here we used the indexing operation $(:,k)$, denoting that the whole column $k$ of the matrix $S$ is taken as a vector - and  the element-wise vector multiplication operation denoted as $x\circ y$ (as well as the normal scalar product as $x^Ty$). Graphical illustration of the eSPA+ decision function geometry in the  in the weighted Euclidean space  - where the dimensions $d$ are weighted with  $\sqrt{W_d}$ - is provided in the Fig.1.
 
 
 \begin{figure}[h!]
        \advance\leftskip-0cm
        \includegraphics[clip,  width=1.0\textwidth]{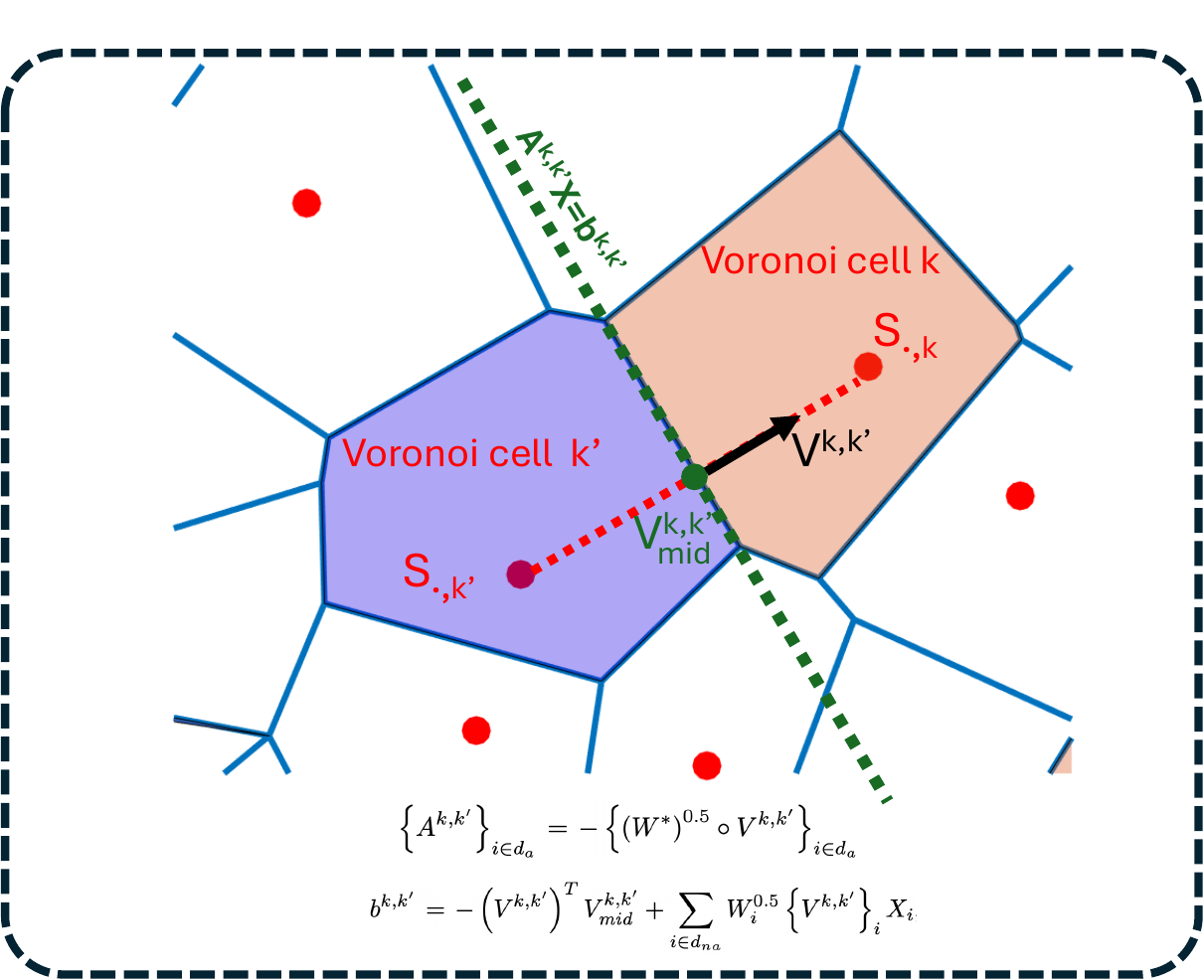}
 \caption{Graphical illustration of the eSPA+ Voronoi cell geometry in the Euclidean spaces dimensions $d$ weighted with  $\sqrt{W_d}$, described by the formulas (\ref{eq:Voronoi_vec},\ref{eq:Voronoi_mid}) and  Theorem 1.}
\end{figure}

 
 Adopting the untargeted attack formulation (\ref{eq:L}-\ref{eq:L_con_ind}), that for any given and fixed feature vector instance $X$, conditions for existence, uniqueness and computational scalability of MAP vector $X^*$  in eSPA are given by the following theorem:
 \begin{theorem}
\label{theorem:thetwo}
Let $X\in\mathbf{R}^D$ is a given and fixed feature vector, $\bar{X}\in\mathbf{R}^D$, $X^*\in\mathbf{R}^D$ be the minimal adversarial perturbation (MAP) of $X$, dist$(X,\bar{X})=\|X-\bar{X}\|_2^2$, $\bP^{\textrm{eSPA+}}_L(\cdot)$ is given by (\ref{eq:bP_espa}), and $ k^* = \arg\min_{k}\sum_{d=1}^DW^*_d\sum_{t=1}^T(X_{d}-\left\{{S^*}\right\}_{dk})^2$. Then, for any $k=1,2\dots,k^*-1,k^*+1,\dots,K$:
\begin{eqnarray}
\label{eq:L_qp}
X^{*,(k)}_{i\in d_a}&=&\arg\min_{\bar{X}_{d_a}}\sum_{i\in{d_a}}(X_i - \bar{X}_i)^2,\\
\label{eq:L_con_ind_espa2}
X^{*,(k)}_{i\in d_{na}}&=&{X}_{i\in d_{na}}.\\
\label{eq:L_qp_con}
\textrm{such that }&&A^{k,k'}{X}^{*,(k)}_{i\in d_a}\leq b^{k,k'},\\
\textrm{with }\left\{A^{k,k'}\right\}_{i\in d_a}&=& -\left\{\left(W^*\right)^{0.5}\circ V^{k,k'}\right\}_{i\in d_a}, \nonumber\\
\textrm{and }b^{k,k'}&=& -\left(V^{k,k'}\right)^TV_{\textrm{mid}}^{k,k'}+\sum_{i\in{d_{na}}}W^{0.5}_i\left\{V^{k,k'}\right\}_iX_i, \quad k'=1,2\dots,k-1,k+1,\dots,K,\nonumber
\end{eqnarray}
with $V^{k,k'}$ and $V_{mid}^{k,k'}$ defined in (\ref{eq:Voronoi_vec}) and (\ref{eq:Voronoi_mid}), respectively. If $\mathcal{J}\subseteq\left\{1,2\dots,k^*-1,k^*+1,\dots,K\right\}$ is a subset of all indices $k$ such that the linear inequality constraints (\ref{eq:L_qp_con}) define a non-empty set, then there exist up to $N\leq\textrm{card}(\mathcal{J})\leq(K-1)$ solutions of the MAP-problem (\ref{eq:L}-\ref{eq:L_con_ind}) (where $\textrm{card}(\cdot)$ measures the number of elements), from a subset of solutions to the problems (\ref{eq:L_qp}-\ref{eq:L_qp_con}), i.e.:
\begin{eqnarray}
\label{eq:L_qp_fin}
J^*&=&\arg\inf_{j\in\mathcal{J}}dist(X,X^{*,(j)}),\nonumber\\
\textrm{and }X^*&=&X^{*,(j)},\quad \textrm{ for all  } j\in J^*
\end{eqnarray}
 Computational cost of the problem scales polynomially in dimension $D$ and linearly in the number $K$ of eSPA Voronoi cells.  
\end{theorem}
{\bf{Proof}}: first, we exploit the observation that finding $X^*$ in (\ref{eq:L}-\ref{eq:L_con_ind})  is equivalent to finding a minimum of the shortest distances between $X$ and $(K-1)$ domains defined by (\ref{eq:bP_espa}). The piecewise-linear boundaries of these $(K-1)$ domains in the subset of controllable/attackable features $d_a$ are defined by the inequality constraints (\ref{eq:L_qp_con}). As follows from the Proposition 3.1 and Theorem 3.1 in \cite{rockafellar70}, constraints (\ref{eq:L_qp_con}) (being a finite intersection of linear half-spaces) constitute a convex polyhedral set for any $k$ \footnote{This can also be directly proven by contradiction, getting use of the definition of the convex function and $W$-weighted squared Euclidean distance $\sum_dW_d\|x_d-y_d\|_2^2$. Interestingly, convexity is given even if some of $W_d\equiv0$. }.   Hence, the corresponding $(K-1)$ optimal solutions $X^{*,(k)}$ are given by the QP solutions with strictly-convex functions (\ref{eq:L_qp}) on the  $(K-1)$ convex sets  (\ref{eq:L_qp_con}). According to the Theorem 2.1.12 in  \cite{rockafellar70}, if these sets are non-empty, then each of these strictly convex problems will have a unique solution. Any such strictly-convex QP with a positive definite Hessian and linear constraints (convex polyhedral set) can be solved in polynomial time using interior-point methods or active-set algorithms that have polynomial iteration bounds. The complexity of these algorithms is typically $\mathcal{O}((D+K)^3 L)$, where $D$ is number of variables, $K$ the number of constraints, and $L$ is the bit-length \cite{boyd04}. Verifying the non-emptiness can be performed with standard rank criteria from the linear algebra, e.g., with the Carver criterium (see Section 3.7.8 in \cite{ilp_theory})$\square$ 

 \subsection{Self-Attacking  eSPA+ (SA-eSPA+) for a data-minimal learning and optimal control}
\label{sec:sa-espa}

The central idea behind the Theorem 1 can be extended also to the other types of optimisation and optimal control problems in AI: for the convex optimisation problems with non-convex constraints  (\ref{eq:L_con}-\ref{eq:L_con_ind})  defined by the AI decision function $\bP_L(\cdot)$), in the case of eSPA+ these non-convex constraints can be decomposed into $K$ sets of convex constraints defined by the boundaries of the Voronoi cells (see Fig.1). Then the overall global problem can be solved selecting an optimum from $K$ convex optimisation problems on convex sets, that have well defined conditions for existence and uniqueness of the obtained solutions.   

This idea also opens a way to creating data-minimal Self-Attacking training methods based on eSPA+ (SA-eSPA+): it is easy to verify that in terms of the probability measure $\Gamma$ , the feature vectors with maximal (in $\Gamma$) entropy  will be positioned at the points $V_{mid}^{k,k'}$, for all combinations of  $k$ and $k'$ between 1 and $K$ (see Fig.1). Such SA-eSPA+ can start with an initial (small) set of $T_{start}$ training pairs  $\tilde{X}_{start}\in\mathcal{R}^{D\times T_{start}}$, $\tilde{\Pi}_{start}\in\mathcal{P}^{M\times T_{start}}$, train the eSPA+ model for this set, compute $V_{mid}^{k,k'}$ and augment the feature set $\tilde{X}$ with these new points. Then, if one has a possibility to determine the respective labels for these new points  $V_{mid}^{k,k'}$ (e.g., conducting additional experiments or performing a model simulation), one can repeat the training of eSPA+ model on this augmented set of features and labels. This procedure is repeated until the maximal allowed number of training data pairs $T$ is reached. This procedure is similar to the one that is deployed in generative Adversarial Networks (GANs) \cite{wang23}, except that the data generator network would be not required for SA-eSPA+, and that in every step the training data is augmented only with the new data instances that are guaranteed to be maximally-entropic - and, hence, least certain in terms of the underlying classifier.   

\begin{figure}[h!]
        \advance\leftskip-2cm
        \includegraphics[clip,  width=1.2\textwidth]{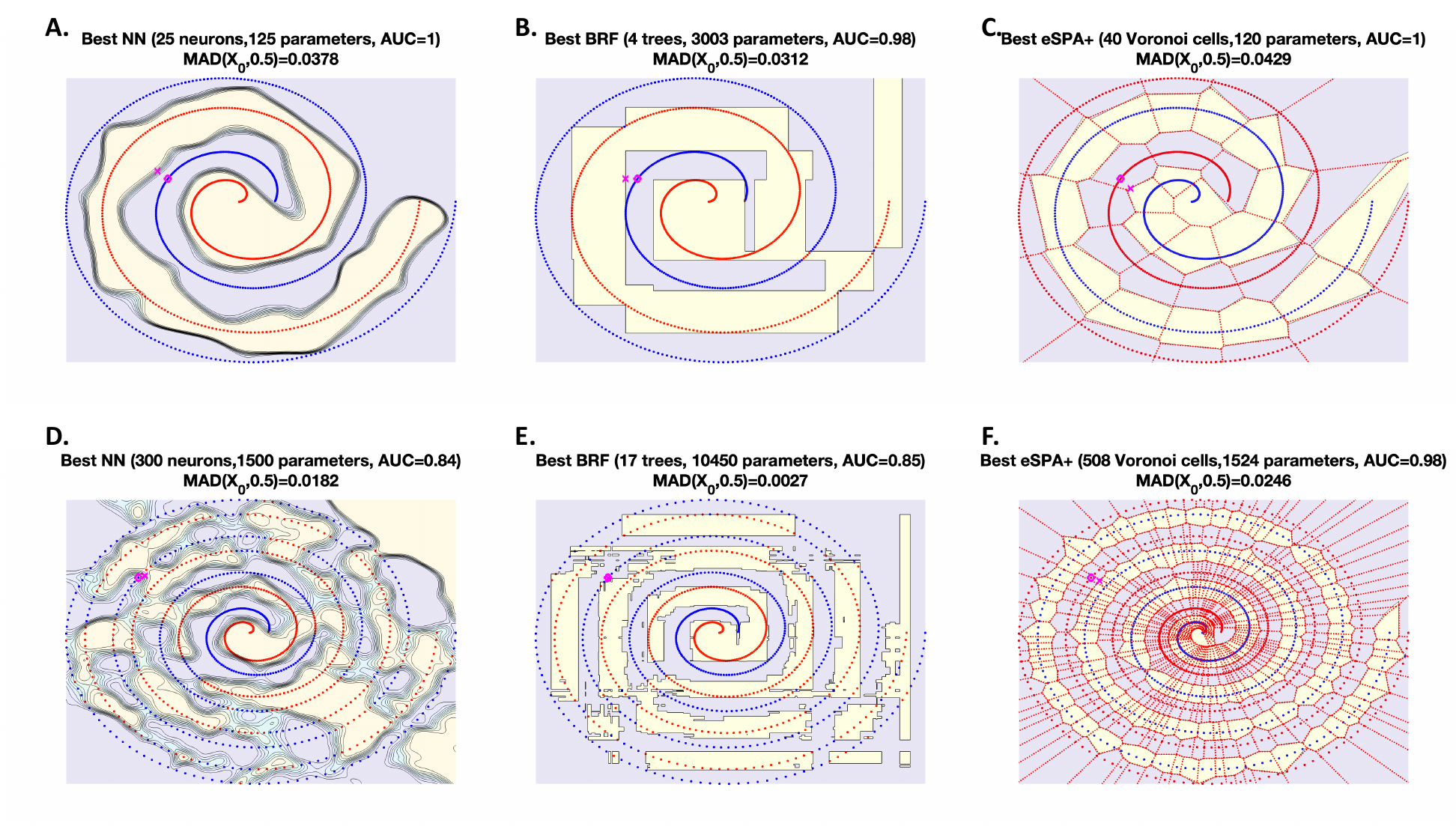}
        \includegraphics[clip,  width=1.2\textwidth]{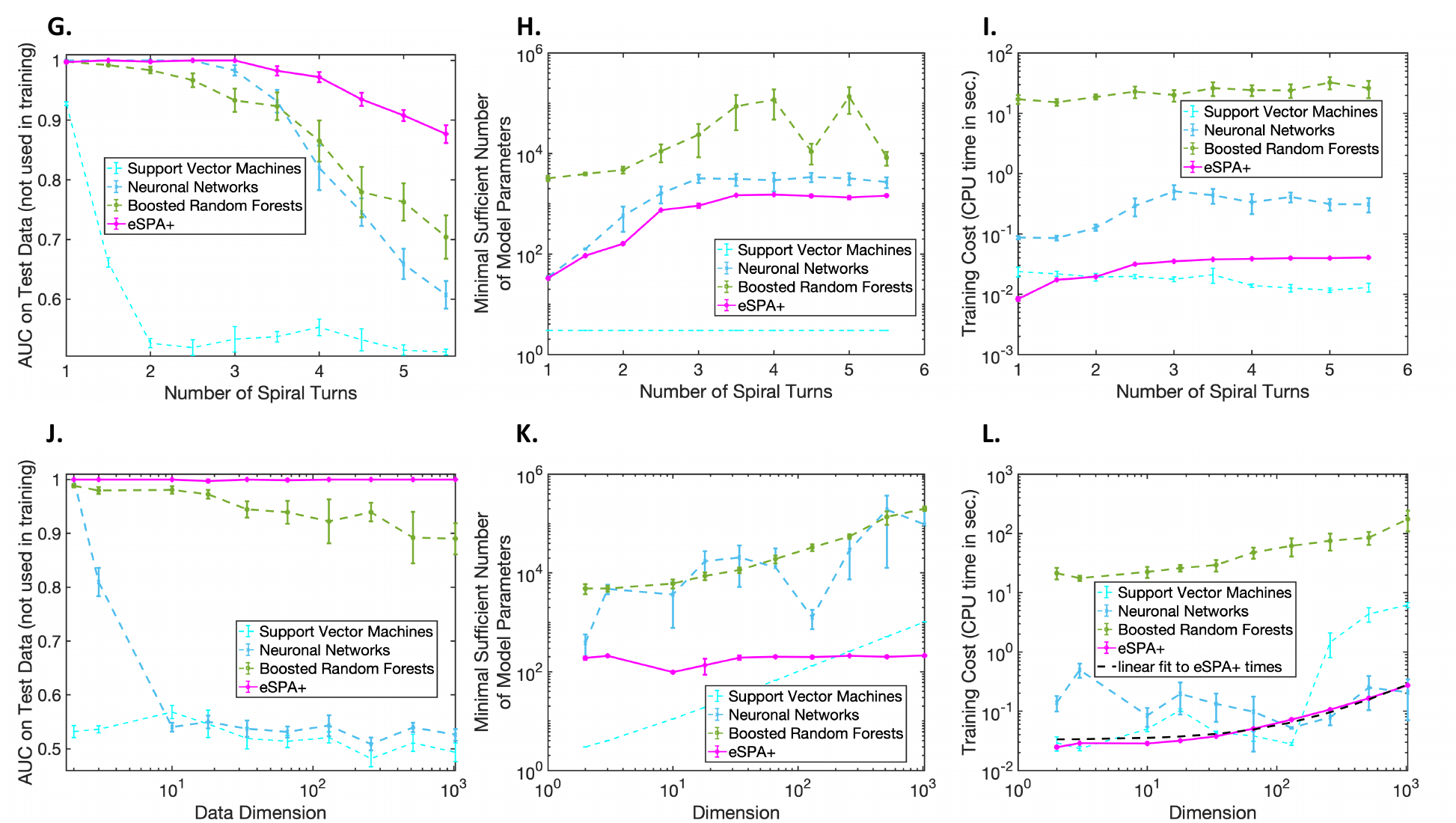}
 \caption{Swiss double-roll spiral and its extensions: MAP/MAD results for a common parameterization with $D=2$ and two full spiral turns (panels A-C), for an extension with four to spiral turns (and all other benchmark parameters left the same as in \cite{swissroll_18}, panels D-F). Panels G-I show comparisons for the changing number of example spirals, and panels J-L for the changing number of non-informative (uniformly-distributed) feature dimensions.}
\end{figure}

\section{Application examples}
\label{sec:results}
In the following, we will demonstrate application of the methods introduced above  for  the synthetic examples and for some real-life applications from biomedicine, comparing among the best performing AI tools configurations  (like the Neuronal Networks, NNs, - finding the best performers among both deep and shallow architectures for various numbers of hidden neurons; among Boosted Random Forest Ensembles, BRF  - finding the best-performing algorithms among XGBoost, RUSBoost and others), TabPFN foundational model for small data learning based on the transformer architectures published very recently \cite{TabPFN}, as well as among eSPA+ and the other AI classifier models like GLM etc. To avoid the software-dependent platform bias (i.e., when some methods are available in a better/faster/robuster software platform implementation), the unified commercial software realization for all of the tools (except of TabPFN, that is available as Python-code) was chosen from within the MATLAB 2024a environment. 
Example in subsection \ref{sec:swiss} presents comparison of different methods for standard AI benchmark Swiss double-roll dataset, subsection \ref{sec:biom} illustrates an application of MAP/MAD to computing optimal therapeutic interventions, and subsections  \ref{sec:worms}-\ref{sec:chemotherapy} showcase the applications of SA-eSPA+ to data-minimal learning in optimisation and optimal control problems. 

\subsection{Swiss double-roll spiral and its extensions}
\label{sec:swiss}
First, we will consider one of the most simple and standard AI classification benchmarks - a two-dimensional Swiss roll with two spirals, making two full turns (each spiral is characterized by a distinctive class affiliation, see Fig 2A-2C), the two spirals are intertwined in each other \cite{swissroll_18}.  This is one of the most basic low-dimensional AI benchmarks, its advantage to the more complex synthetic problems is that it is very easily extendable to become more complex - and to cover more advanced scenarios. In contrast, more complex examples are much more difficult to be boiled down to the clear and insightful level of the Swiss double-roll in 2D.  For example, Swiss double-roll  is easily extendable to multiple dimensions:  to illustrate applications of nonlinear dimensionality reduction approaches, one adds additional non-informative dimensions (usually with uniform data distribution) to the two dimensions in which the two spirals are distinguishable \cite{bengio03}.   

We start with the most common 2D benchmark parameter setting, that has two full turns for each of the spirals, and uses 512 random data points for training and another 512 for validation and testing (a $50\%-25\%-25\%$ cross-validation split) \cite{swissroll_18}. Panels A-C of Fig.2 show the results for this setting for the best performers among NNs (Fig. 2A), among boosted random forests (BRF, Fig. 2B), and for eSPA (Fig. 2C).  The number of parameters for each of the models is determined as the minimal model descriptor length that is sufficient for the model to achieve the maximal AUC on the test data: e.g., for neuronal networks (NNs), we scramble over different architectures and different numbers of neurons, selecting the model that is simplest and yet is able to achieve the maximal test AUC that is observed on the whole set of architectures.   Magenta circles denote the randomly selected point $X$ that has to be subject to an adversarial attack and MAD/MAP computation. Magenta crosses represent the computed endpoints $X^*$ of MAP for each of the methods. To compute $X^*$ for eSPA+, we deploy the Theorem 1. For other methods (NN, boosted Random Forests), we use the direct numerical solution of the problem (1-3), implemented with the standard  \emph{fmincon}-function of MATLAB for constrained mininimization.  Panels D-F of Fig.2 illustrate the challenge of extending this simple benchmark with the increasing number of spiral turns - by keeping all other example parameters constant, including the size of the training data statistics that stays 512 as in the original benchmark from  \cite{swissroll_18}. As can be seen from Fig. 2D-2F, with a growing number of spiral turns, increasing non-convexity of  $\bP_L(\cdot)$ (see contour-lines of Fig.2A-2F)  results in the shorter adversarial perturbations and smaller MADs.
 Panels Fig. 2G-2I show dependence of  classifier performances for a changing number of spiral turns in the standard example, whereas panels Fig. 2J-2L  illustrate how the performance of AI classifiers change with the added number of the non-informative uniformly distributed dimensions.

\subsection{Using MAP for a patient-specific computing of the unique minimal risk-minimising therapeutic interventions}
\label{sec:biom}
\begin{figure}[h!]
        \advance\leftskip-0cm
        \includegraphics[clip,  width=1\textwidth]{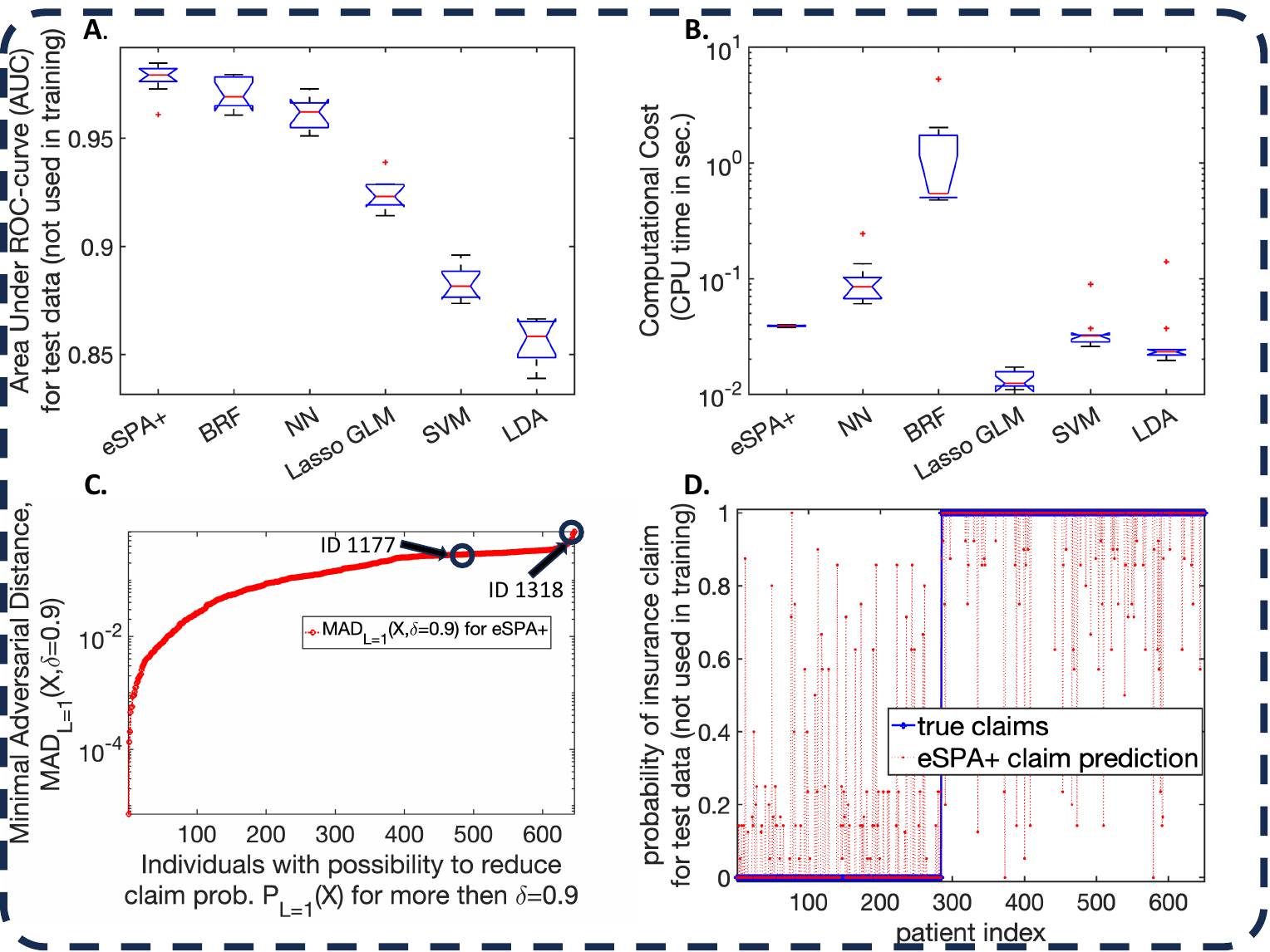}
        \includegraphics[clip,  width=1\textwidth]{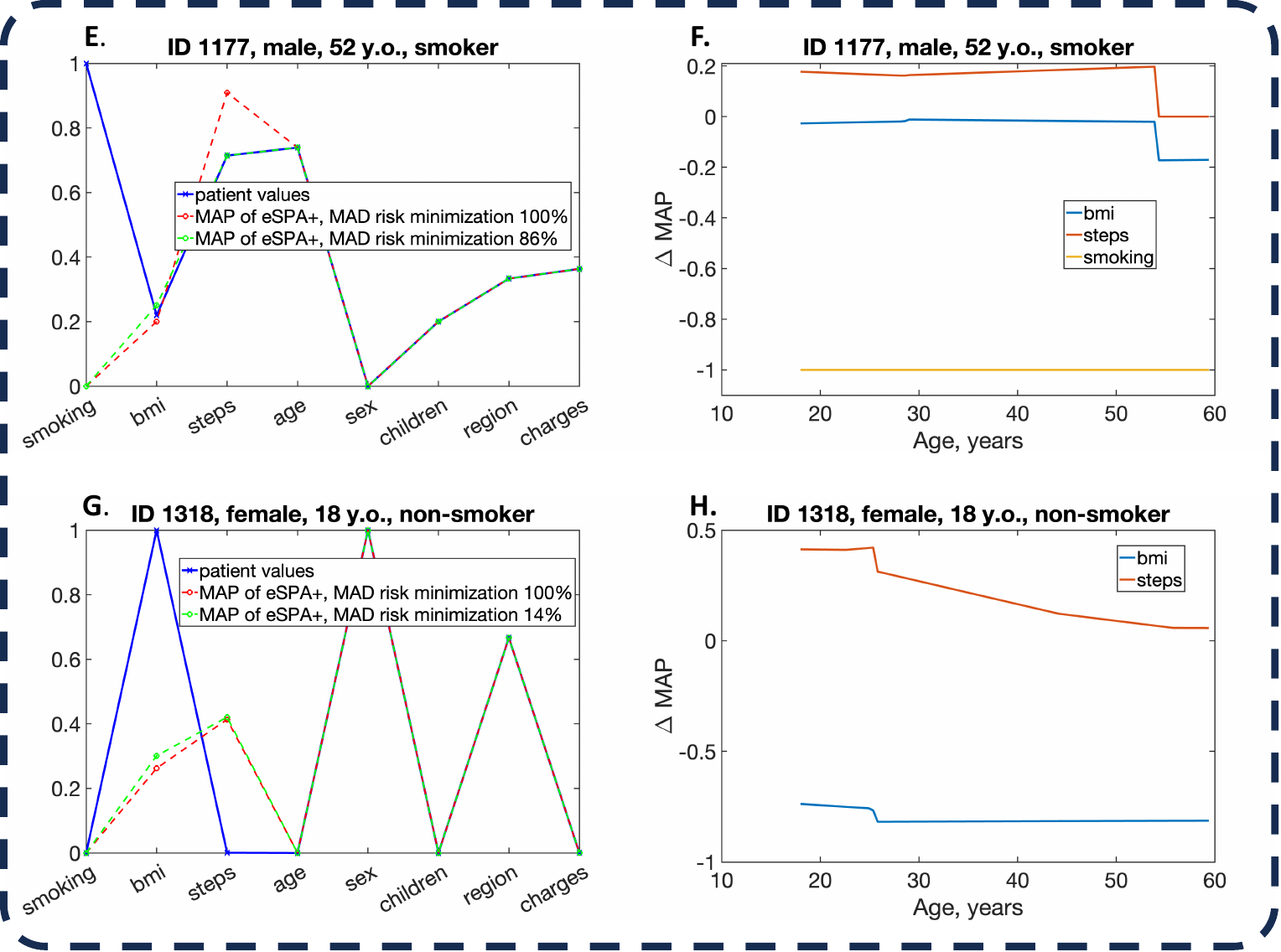}
 \caption{Computation of MAP and MAD for the biomedical data analysis. Panels A and B represent the standard statistical boxplots, where the red line denotes the median and the blue bars the sizes of quantiles. }
\end{figure}

Next, in Fig. 3  we see the results for an application to the medical insurance claim database. As the controllable/attackable features we consider the smoking status, the Body-Mass-Index (BMI) and the average number of steps the person is doing per day.  A link to this openly-accessible data source is provided in the Section "Availability of data and material". As can be seen from the panels A-D of Fig. 3, eSPA is providing a superior prediction of cases on a test data (i.e., patient cases not used in training and validation). It achieves the value of AUC=0.98.   Computed MAPs (in panels E-H of Figs. 3)  allow an interpretable - and mathematically-guaranteed minimal (due to the above Theorem 1) therapeutic intervention that would have allowed a reduction of risk with a predefined threshold  $\delta=0.9$ of the insurance claim risk. According to these results, approximately three-fourths of the insurance claiming individuals could have avoided claims through an appropriate adjustment of BMI, of the number of daily steps and of their smoking status.  Moreover, obtained  results allow interpretable insights into the age-dependence of these optimal interventions (see Figs. 3F and 3H).     
         
\subsection{Self-attacking data-minimal learning in C. elegans worms longevity model}
\label{sec:worms}
\begin{figure}[h!]
        \advance\leftskip-1.5cm
        \includegraphics[clip,  width=1.2\textwidth]{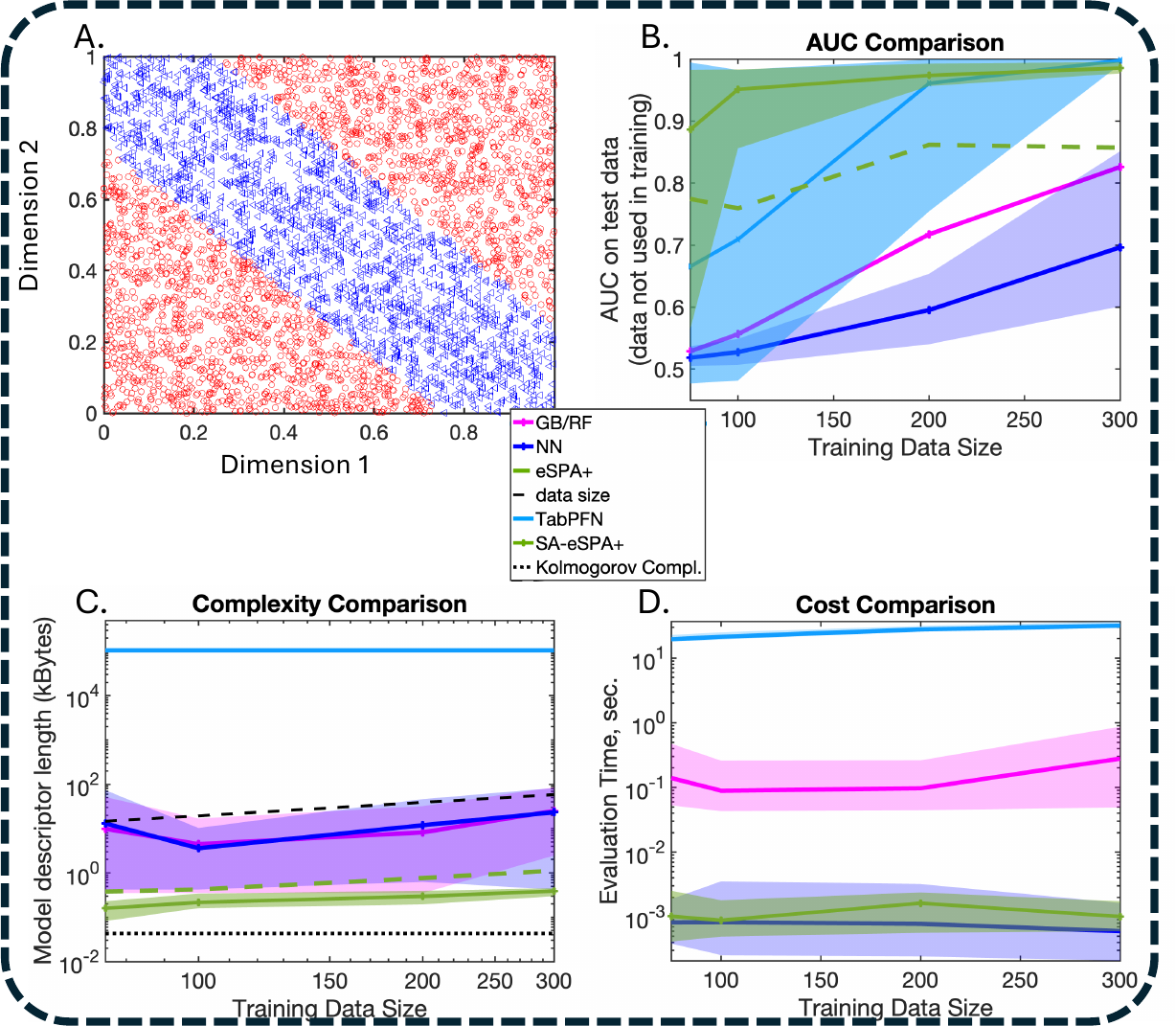}
 \caption{Self-Attacking Entropic Learning (SA) in comparison to existent AI/ML tools for a worms longevity data example, for D=50. Panels B-D feature the means and the 95\% confidence intervals computed  from the standard cross-validation procedure. }
\end{figure}

 Lan et al. \cite{lan19} have shown how a certain type of
worm (Caenorhabditis elegans) can sometimes exhibit a significantly longer
life span than average, thanks to the orchestrated expression of two specific genes - mTOR and IIS. 
The fact that this increased longevity is induced
solely by the co-expression of a single pair of genes makes the distinction between the long-living and normal worms a challenging task: the worms can be
correctly classified only according to the two longevity-relevant genes, but
this significant pair of features (Dimension1 and Dimension2 in Fig.4A) is hidden among thousands of other genes,
that do not effectively contribute predictive information in this context. In
this example, we thus aim at correctly discerning the sparse learning rule
that pinpoints only the relevant pair of feature dimensions,  allowing an optimal distinction
 of the long-living worms (blue triangles in Fig.4A) from the ones with a normal life
span (red circles in Fig.4A).  This data benchmark is one of the most generic and simplest examples of a
classification problem with a data statistics that is not separable by a linear manifold. The feature vectors 
of the long-living worms are here diagonally-sandwiched between the feature vectors
coming from the regular worms. The
other $D-2$ dimensions are uniformly distributed data samples with the
same variance, where the two classes of worms are indistinguishable.  
An
important feature of this very generic model is the fact that it can be tuned
to include an arbitrary number of worms (i.e.,observations) and genes (i.e., feature dimensions). This allows us to provide, in the validation part, both
an appropriate scaling analysis of various learning algorithms for a variety of small data
problems, and a comparison of different methods for various combinations
of data statistics $T$ and feature dimensions $D$.

Another important feature of this benchmark is that one can compute the exact Kolmogorov complexity for this problem, i.e., the minimal descriptor length of the simplest learning model 
 that is sufficient to provide a rule that distinguishes between the two groups of worms. It is straightforward to validate that such a minimal model would require not more then 11 tuneable parameter (6 to encode the centroids of 3 groups in 2D, 3 to encode their labels, and 2 parameters to encode the indices of the relevant feature dimensions). This can be used to compare and to rank different models in terms of their complexity, measured as the model descriptor length in kBytes - and to compare the complexity of the trained model both to the Kolmogorov complexity and to the complexity of the available training data. This is a very useful comparison, since it would allow to find out whether the different AI tools really 'learn' the underlying  rule - or if they rather 'memorise' the training data.    

Fig.4B shows comparison of Area-Under-Curve (AUC) measure of the prediction performance for D=50, evaluated on the test data that was not available for model training. To control the statistical-significance of the provided AUC values we use the same standard cross-validation procedure for all of the deployed methods: as described above, we subdivide the data in 20 randomly chosen  training, validation and testing subsets, plotting the mean and the 95\% confidence intervals for all of the obtained AUCs. SA-eSPA+ is implemented according to the procedure described in \ref{sec:sa-espa}, initialised with $T_{start}=60$, and with the maximal allowed number of training points being $T$ (the x-axis of the Fig.4B). SA-eSPA+ statistically-significantly outperformed all of the other considered methods, with the closest competitor being the foundational TabPFN transformer model \cite{TabPFN}. In terms of model complexity, SA-eSPA+ is also outperforming other considered models, having the complexity that is closest to the exact Kolmogorov complexity of the underlying true model - whereas the descriptor length of other competitors rather follows the complexity of the training data, indicating that these models rather 'memorise' and not 'learn' (Fig.4C).
 TabPFN is a very large transformer model,  with around 28 millions of tuneable parameters \cite{TabPFN}. Hence, its complexity is around 5 orders of magnitude larger then the complexity of SA-eSPA+. Accordingly, as can be seen from Fig.4D, cost of evaluating TabPFN on a single feature vector instance (measured in CPU seconds) is around 4 to 5 orders of magnitude larger then the cost of the SA-eSPA+ evaluation.

 \subsection{Self-attacking data-minimal learning of optimal controls in Lorenz-63 dynamical system}
 \label{sec:lorenz}

 Next, we illustrate the application of SA-eSPA+ to a direct learning of optimal controls in the simple chaotic  forced dissipative hydrodynamic flow system called Lorenz-63 \cite{lorenz63} in the standard parameter setting, but with the external forcing:
 \begin{eqnarray}
\dot{x}(t) &=& 10(y(t) - x(t)) +A_x(t)u_x(t), \nonumber\\
\dot{y}(t) &=& x(t)(28 - z(t)) - y(t) +  A_y(t)u_y(t),\nonumber\\
\dot{z}(t) &=& x(t)y(t) - \frac{8}{3}z(t) + A_z(t)u_z(t).
\end{eqnarray}
  Adding the forcing terms $A_x(t)u_x(t), A_y(t)u_y(t), A_z(t)u_z(t)$  to the x, y, and z derivative components of this three-dimensional dynamical system, we create a control problem were one can switch on-and-off control terms in the right-hand-sides of the three respective ordinary differential equations, with  $A_x(t), A_y(t), A_z(t)$  being allowed to attain only values 1 (when control in the respective component is on), and 0 (when control in the respective component is off). The maximal simulation time is set to $T_{control}$, taking values $4.0$, $5.0$ and $6.0$ (see Fig 5). The control is designed such that every control term  $A_x(t)u_x(t)$ and $A_y(t)u_y(t)$ is allowed to take constant values $(u_x(t_1), u_y(t_1))$, $(u_x(t_2), u_y(t_2))$ and $(u_x(t_3), u_y(t_3))$ between 0 and 100 at fixed times  $t_1=0$,  $t_2=1$ and  $t_3=2$ for a duration $\delta t=0.1$ - and only zero values outside of these time intervals (as a so-called 'bang-bang-control').  $u_z(t)$ is allowed to take constant values $u_z(t_1)$, $u_z(t_2)$,  $u_z(t_3)$  between 0 and 100 in these time intervals - but the $A_z(t)$ is set to zero for all $t$. 
 
 The information about $A(t)$, about the equation parameters and solution trajectories  for $0<t<T_{control}$ is not provided to the SA-eSPA+: it is done to emulate the situation when the underlying dynamical system is a 'black box', and be able to check if the method will be able to distinguish between the important, the explicitly-unimportant (like $u_z(t)$) and implicitly-unimportant control dimensions (e.g., some subset of control dimensions along $x$ and $y$ directions with the non-zero value, but without explicit importance for the particular control optimality conditions of interest). We initialise  the system at $t=0$ very  close to the fixpoint of the right attractor lobe, such that for the selected complete simulation time $t_{final}=3$ it would remain very close to this fixpoint (see the blue curve in Fig.5C). The optimal control task is to find the $l_2$-minimal 9-dimensional vector $X=(u_x(t_1), u_y(t_1), u_z(t_1),u_x(t_2), u_y(t_2), u_z(t_2),u_x(t_3), u_y(t_3), u_z(t_3))$ of control variables, such that the system starting at $t=0$ at the fixed initial condition very  close to the fixpoint of the right attractor lobe ends-up being in the left attractor lobe at $T_{control}$. For every control vector $X\in\mathcal{R}^9$, we can use the Runge-Kutta-solver to simulate the forced Lorenz-63 under these conditions. If the control task was successful and the Lorenz-63 system ended-up in the left attractor lobe for a given control vector $X$, we can assign a label $1$ to this vector and assign zero if this control task was unsuccessful. Thus, we can repeat this procedure vor various (random) values of $X$ that satisfy the above-mentioned control amplitude constraints - and collect the respective information into the feature matrix $\tilde{X}$ and label probability matrix $\tilde{\Pi}$, transforming this control problem to     
a supervised binary classifier learning problem.

Please note that since (a) the system is emulated as 'black box', with no trajectory information for $0<t<T_{control}$ provided to the learning agent, and (b)  $T_{control}$ are selected very large and out of prediction range for common learning tools (since for selected $T_{control}$ and dependent of the initial position, system has a chance to cross the boundary between the attractor wings several times),  methods like reinforcement learning and physics-informed NNs (PINNs) - that try to predict the next step out of the time series history and for this system work reliably only for $T_{control}<0.1$ -can not be applied here to solve this optimal control problem. Instead, as a benchmark for SA-eSPA+, we will use here the sensitivity derivative Monte Carlo method (SDMC) for optimal control problems, that is guaranteed to find the globally-optimally solution when the sample size $T$ asymptotically approaches infinity \cite{montecarlo03}.     

Results of the numerical tests  for $T_{control}$ taking values $4.0$, $5.0$ and $6.0$ are summarised in Fig.5. First, independent of the selected value of  $T_{control}$, inspection of the inferred feature importance weights $W$ defined in (\ref{eq:L_tilde}) (see Fig.5A, Fig.5D, Fig.5G) shows that SA-eSPA+ correctly finds that the z-components of control are explicitly-unimportant and can be switched off.  Moreover , it identifies that the x-components are also not relevant for finding the optimal control and, hence, can be switched off too. Fig.5B, Fig.5E and Fig.5H show the comparison of minimal control energies $\sum_{k=1}^3\sum_{d=x,y,z}u_d(t_k)^2$ found by the Monte Carlo method SDMC and SA-eSPA+ as functions of the number of maximally-allowed system simulations. SA-eSPA+ requires around $10^4$ less systems simulations to find the optimal (l-2 energy minimal) optimal controls. Moreover,  Fig.5C, Fig.5F and Fig.5I show that  SA-eSPA+ systematically finds almost $40\%$ more optimal controls than SDMC. Moreover, in contrast to SDMC, SA-eSPA+ reveals that with the increasing value of $T_{control}$ - i.e., with the system given more time to achieve the left attractor lobe from its initial position - the energy of the minimal control can be reduced accordingly.     
\label{sec:lorenz}
\begin{figure}[h!]
        \advance\leftskip-1.5cm
        \includegraphics[clip,  width=1.2\textwidth]{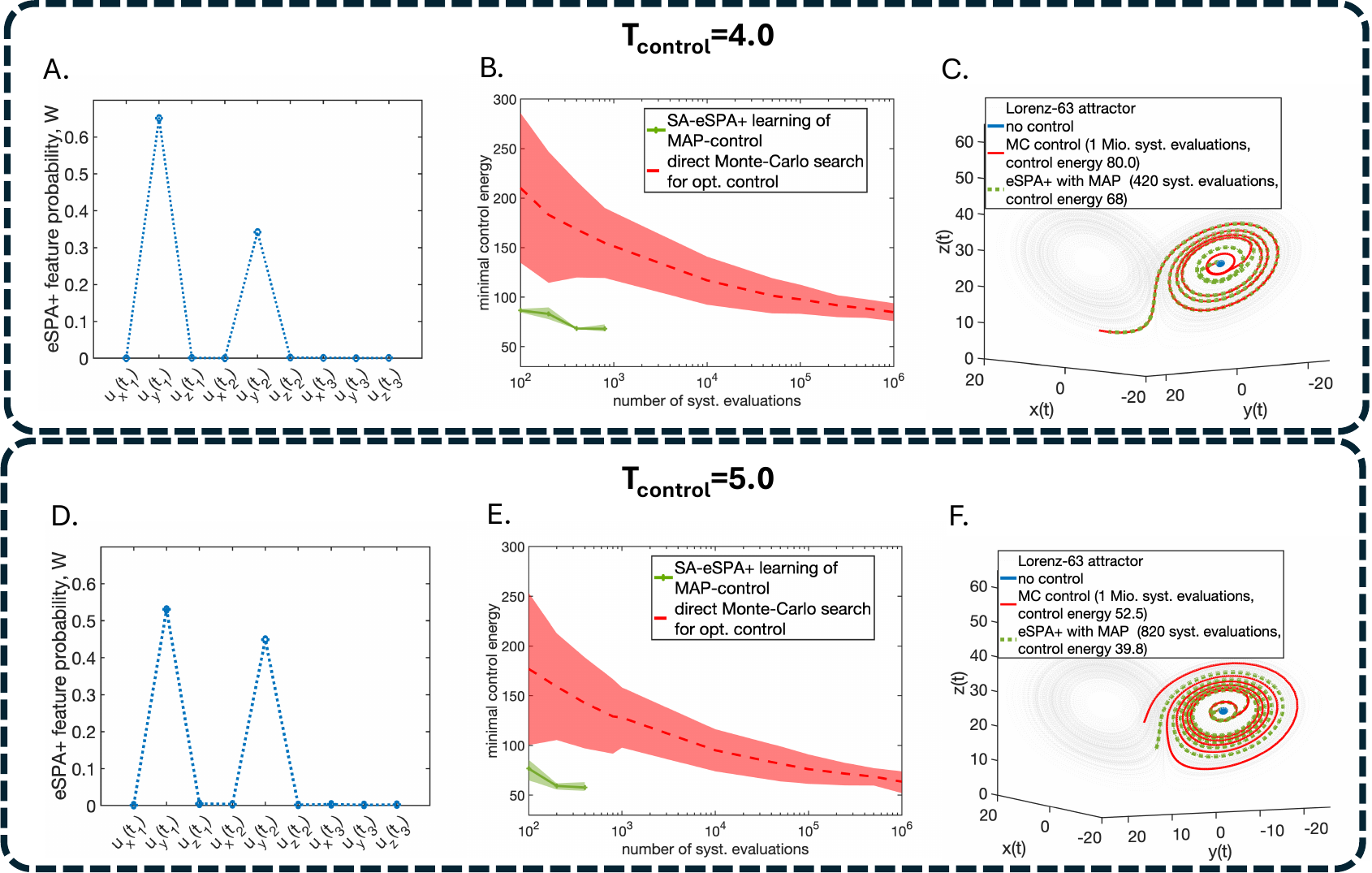}
        \includegraphics[clip,  width=1.2\textwidth]{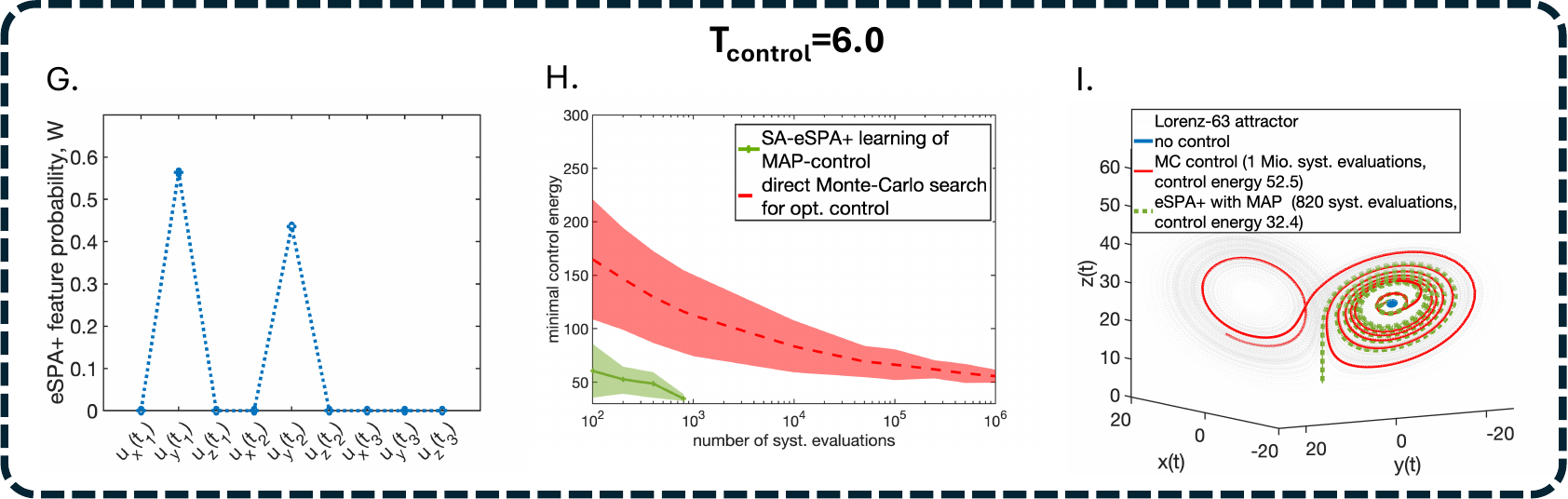}
 \caption{Application of SA-eSPA+ for a direct learning of optimal control in chaotic blackbox-systems, illustrated on a  Lorenz’63-example. Panels B,E, H feature the means and the 95\% confidence intervals computed from the standard cross-validation procedure. MAP-control in the legends of panels  B, E and H refers to the computation of MAP according to Theorem 1 above, the word "control" here just underlines the meaning of the MAP that is found - since it aims at providing the optimal \emph{control} of the underlying system in this application.}
\end{figure}

\subsection{Self-attacking data-minimal learning of optimal drug administration schedules in Acute lymphoblastic leukemia (ALL) chemotherapy for children}
\label{sec:chemotherapy}

Previous example showcases an application to a relatively-simple dynamical system, with three coupled ordinary differential equations and a low simulation cost. 
 To illustrate the power of the proposed methodology for much more complex systems, we next consider the problem of learning the optimal drug administration schedule for children with  
  Acute lymphoblastic leukemia (ALL). Selecting the appropriate therapy and drug administration plan depends on multiple patient-specific characteristics like age, weight, sex, clinical history etc., and is regulated by the clinical guidelines \cite{leukemia20}.  In the following, to compare the drug administration plans and their outcomes for the guideline-based and the AI-based approaches,  we will use an extension of the pharmacokinetic model for cancer therapy \cite{pkpc09} available in the SimBio-Toolbox of MATLAB, coupling it to the pharmacokinetic models for further organs (kidney and liver), and adjusting the parameter setting to allow for modelling ALL therapy effects for children. Resulting model has thousand of coupled equations, and its computer simulation  is around $5\cdot10^3$ times computationally more expensive then the Lorenz-63 model from the previous example. 
  
 As patient settings for model initialisation, we select age being 5.0 years, weight equal 20 kg and the initial total amount of cancer tissue as 1.0 gram. We will consider the three-times-daily drug administration strategy plan, introducing as control variables the drug amounts, the starting times and the numbers of repetitions for each of the three daily sessions, resulting in the 9-dimensional control vector. We also consider constraints on the maximal single drug doses, on the maximal duration of the full therapy (28 days) and on the maximally allowed drug toxicity threshold in liver.   Drug administration  is considered to be successful if non of these constraints is violated during the simulated therapy at any time - and when for a given control vector $X$ on day 28 since the start of the therapy, the total weight of the cancer cells is reduced to less the $10^{-4}$ grams. In this case the label corresponding to this control vector $X$ is set to 1, otherwise to 0. First we, compare the complexity and the training costs of NNs, gradient boosted Random Forests (GB/RFs) and SA-eSPA+ for the obtained supervised classification problem. As can be seen from the left and central panels of the Fig.6A, SA-eSPA+ outperforms the NNs and  GB/RFs in terms of training costs. Complexity of SA-eSPA+ remain roughly unchanged with the training data size, whereas the complexities of NNs and  GB/RFs  follow the size of the training data set - indicating that both NNs and  GB/RFs tend to rather 'memorise' and not 'learn' the underlying problem.       
 
 Next, we consider the optimal control task for this problem, searching for a therapy that is successful (has a label 1) and such that the overall amount of administered drug is minimal. 
 Right panel of Fig.6A shows the results of comparing the minimal sufficient total drug amounts as functions of the number of system simulations, obtained for the optimal therapy plans learned with  SDMC \cite{montecarlo03} and with the SA-eSPA+.  SA-eSPA+ requires around 500 times less system simulations to learn the optimal therapy plan, and can be computed in around 20 minutes on a usual laptop PC - whereas SDMC requires around 22 hours to obtain the control of the similar quality on the same computer. Fig.6B shows comparisons of the time series of drug concentrations in different organs, obtained for the guideline-based plan  \cite{leukemia20} (left column of Fig6B), together with the time series obtained from the optimal plans learned with the SDMC \cite{montecarlo03} (central column of Fig6B), and with the SA-eSPA+ (right column of Fig6B). SA-eSPA+ finds a successful therapy plan with the lowest total amount of the administered drug (with 4.4 times smaller drug amount than the one from the guidelines) and the lowest drug concentrations in liver (around $75\%$ lower than the toxicity threshold) . 
\begin{figure}[h!]
        \advance\leftskip-1.5cm
        \includegraphics[clip,  width=1.2\textwidth]{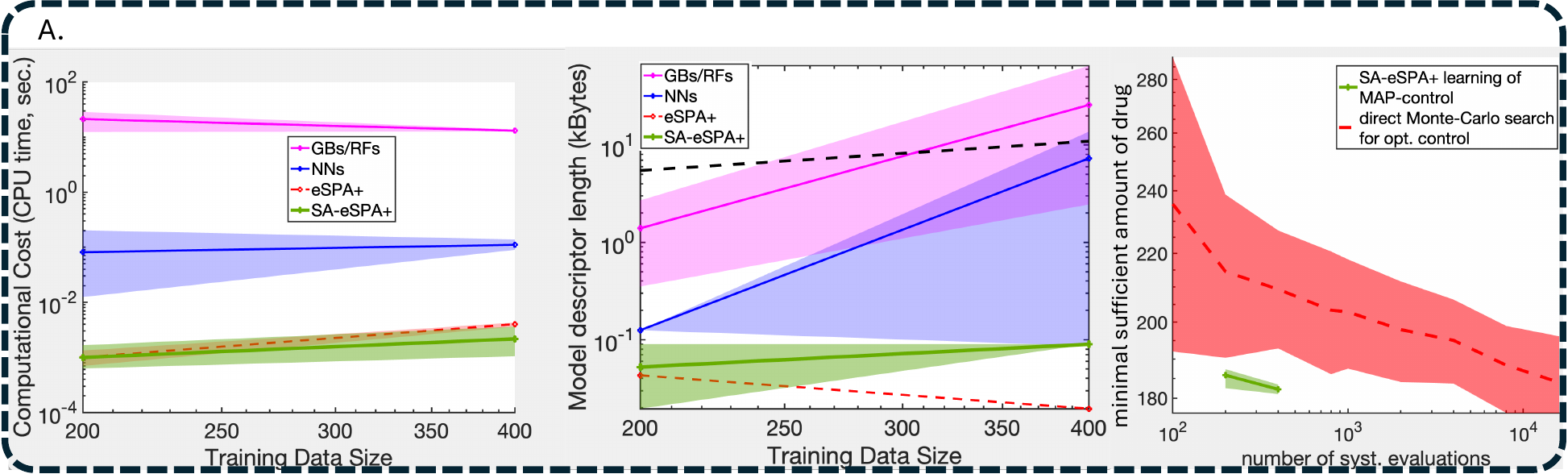}
        \includegraphics[clip,  width=1.2\textwidth]{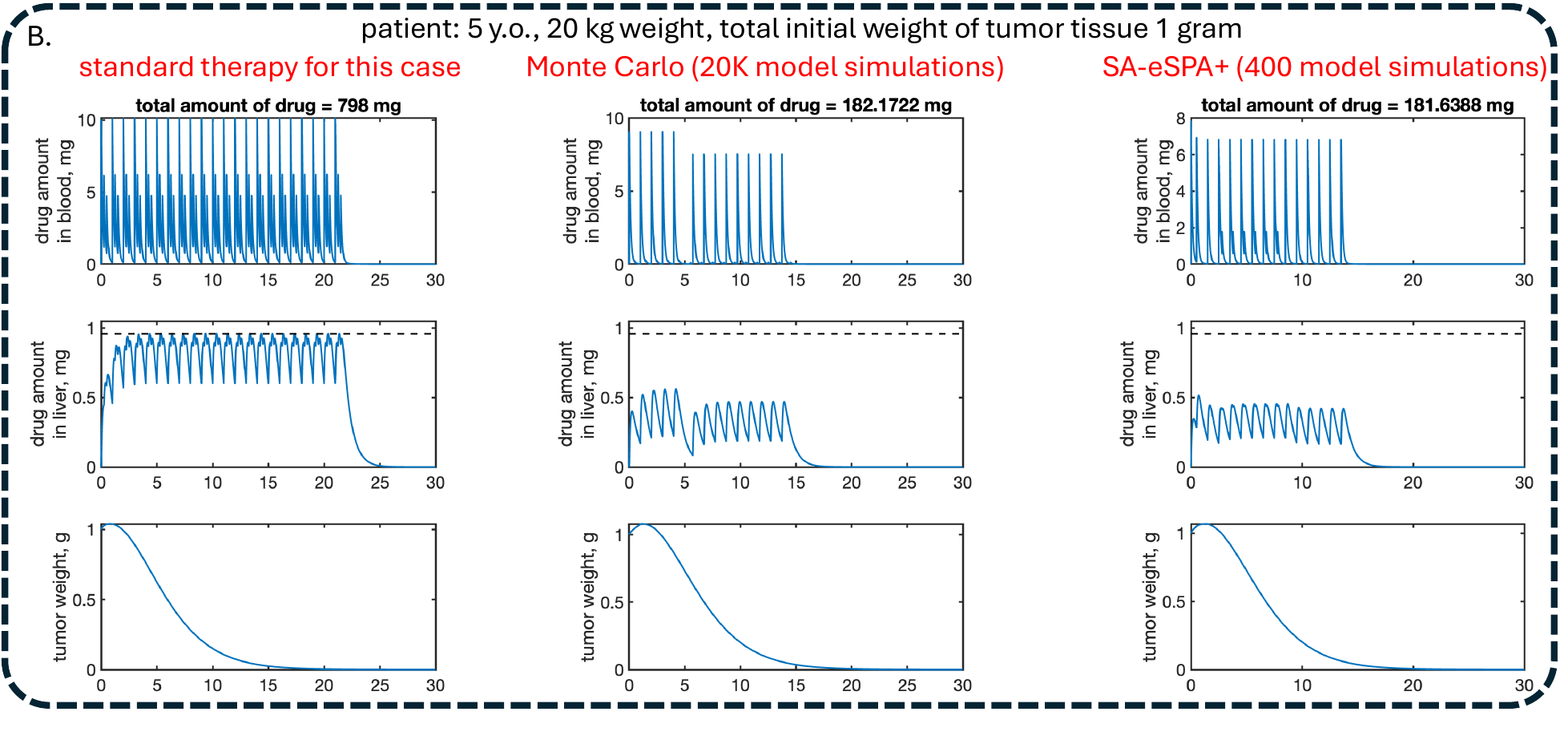}
 \caption{Application of SA-eSPA+ to a direct learning of  optimal patient-specific chemotherapy plan for children with Acute lymphoblastic leukemia (ALL) from a 
minimal number of pharmacokinetic/pharmacodynamic model simulations. Panel A features the means and the 95\% confidence intervals computed  from the standard cross-validation procedure. MAP-control in the legend of panel A refers to the computation of MAP according to Theorem 1 above, the word "control" here just underlines the meaning of the MAP that is found - since it aims at providing the optimal \emph{control} of the underlying system in this application.
}
\end{figure}

\section{Discussion}
\label{sec:discuss}
This paper presented an entropic learning approach to a broad class of AI problems, including AI resilience and robustness with respect to adversarial attacks, to a problem of data-efficient learning, and to a problem of data-minimal learning of optimal controls in complex systems. In contrast to the existent tools, presented approach allows construction of mathematically-justified methods that have cheaply-verifiable conditions for the existence and uniqueness of the adversarial problem solutions (see Theorem 1 in \ref{sec:espa_attack}). As explained in \ref{sec:sa-espa}, the central idea behind this Theorem can be applied to construct very efficient  data-minimal learning algorithms like the Self-Attacking entropy-optimal Sparse Probabilistic Approximation (SA-eSPA+), and can be used for a direct data-minimal learning of optimal controls in complex systems. Data-minimality of the introduced approach becomes especially relevant when dealing with fields where generating a single new training data instance requires an extensive computation or an expensive experiment. As demonstrated on various application examples in \ref{sec:results}, SA-eSPA+ outperforms all of the considered common tools in terms of performance, and results in learning models that require several orders of magnitude less data for training. Comparisons of resulting model complexities (measured in terms of the model descriptor lengths) presented in examples \ref{sec:swiss}, \ref{sec:worms} and \ref{sec:chemotherapy} reveal  that the complexity of trained common tools tend to follow the complexity of the training data, indicating that these tools rather 'memorise' then 'learn'. On the same examples, SA-eSPA+ exhibited the smallest complexity that was orders-of-magnitude smaller then the complexity of other tools, and was practically not changing with the change of the training data size.

Summarising, presented results reveal usefulness of the proposed approach for a broad class of AI problems. Algorithms like SA-eSPA+ based on this approach open new - and yet unavailable - possibilities for AI-driven explorative analysis and control of complex systems in various disciplines.

\bmhead{Availability of data and code} Medical data used in Figs. 2 and 3 are available at \url{https://www.kaggle.com/datasets/easonlai/sample-insurance-claim-prediction-dataset} and \url{https://www.kaggle.com/datasets/andrewmvd/heart-failure-clinical-data}, respectively. Code reproducing the reported results can be provided upon a reasonable request.

\bmhead{Acknowledgement} The author would like to express his very big thanks to the Editor - Rupert Klein - for a thorough edit and for providing numerous very helpful comments. Further, I would like to thank Susanne Gerber (JGU Medicine), Davide Bassetti (RPTU), Georg Gottwald (U Sydney)  and the other participants of the Oberwolfach Mini-Workshop  "Mathematics of Entropic AI in the Natural Sciences", who provided a lot of very helpful feedback.
 This work was funded by the EU Horizons project $AI4LUNGS$ (Grant Agreement No. 101080756).

\bibliography{MAD}

@book{boyd04,
  author = {Boyd, Stephen and Vandenberghe, Lieven},
   description = {CiteULike},
  howpublished = {Hardcover},
  isbn = {0521833787},
  month = {March},
  publisher = {{Cambridge University Press}},
  title = {Convex Optimization},
  url = {http://www.amazon.com/exec/obidos/redirect?tag=citeulike-20\&path=ASIN/0521833787},
  year = 2004
}

@book{rockafellar70,
  address = {Princeton, N. J.},
  author = {Rockafellar, R. Tyrrell},
  callnumber = {UniM Maths 516.08 R59},
  description = {robotica-bib},
  notes = {A SRL reference.},
  publisher = {Princeton University Press},
  series = {Princeton Mathematical Series},
  title = {Convex analysis},
  year = 1970
}

@article{pkpc09,
	author = {Koch, Gilbert and Walz, Antje and Lahu, Gezim and Schropp, Johannes},
	date = {2009/04/01},
	doi = {10.1007/s10928-009-9117-9},
	id = {Koch2009},
	isbn = {1573-8744},
	journal = {Journal of Pharmacokinetics and Pharmacodynamics},
	number = {2},
	pages = {179--197},
	title = {Modeling of tumor growth and anticancer effects of combination therapy},
	url = {https://doi.org/10.1007/s10928-009-9117-9},
	volume = {36},
	year = {2009}}

@article{leukemia20,
	address = {Huntington NY, USA},
	author = {Patrick Brown and Hiroto Inaba and Colleen Annesley and Jill Beck and Susan Colace and Mari Dallas and Kenneth DeSantes and Kara Kelly and Carrie Kitko and Norman Lacayo and Nicole Larrier and Luke Maese and Kris Mahadeo and Ronica Nanda and Valentina Nardi and Vilmarie Rodriguez and Jenna Rossoff and Laura Schuettpelz and Lewis Silverman and Jessica Sun and Weili Sun and David Teachey and Victor Wong and Gregory Yanik and Alyse Johnson-Chilla and Ndiya Ogba},
	doi = {10.6004/jnccn.2020.0001},
	journal = {Journal of the National Comprehensive Cancer Network J Natl Compr Canc Netw},
	number = {1},
	pages = {81 - 112},
	publisher = {National Comprehensive Cancer Network},
	title = {Pediatric Acute Lymphoblastic Leukemia, Version 2.2020, NCCN Clinical Practice Guidelines in Oncology},
	url = {https://jnccn.org/view/journals/jnccn/18/1/article-p81.xml},
	volume = {18},
	year = {2020}}

@article{montecarlo03,
	author = {Cao, Yanzhao and Hussaini, M. Y. and Zang, T. A.},
	date = {2003/12/01},
	doi = {10.1023/A:1026079021836},
	id = {Cao2003},
	isbn = {1573-2894},
	journal = {Computational Optimization and Applications},
	number = {3},
	pages = {219--230},
	title = {An Efficient Monte Carlo Method for Optimal Control Problems with Uncertainty},
	url = {https://doi.org/10.1023/A:1026079021836},
	volume = {26},
	year = {2003}}

@article{lorenz63,
	address = {Boston MA, USA},
	author = {Edward N. Lorenz},
	doi = {10.1175/1520-0469(1963)020<0130:DNF>2.0.CO;2},
	journal = {Journal of Atmospheric Sciences},
	number = {2},
	pages = {130 - 141},
	publisher = {American Meteorological Society},
	title = {Deterministic Nonperiodic Flow},
	url = {https://journals.ametsoc.org/view/journals/atsc/20/2/1520-0469_1963_020_0130_dnf_2_0_co_2.xml},
	volume = {20},
	year = {1963}}

@article{lan19,
	author = {Jianfeng Lan and Jarod A. Rollins and Xiao Zang and Di Wu and Lina Zou and Zi Wang and Chang Ye and Zixing Wu and Pankaj Kapahi and Aric N. Rogers and Di Chen},
	doi = {https://doi.org/10.1016/j.celrep.2019.06.078},
	issn = {2211-1247},
	journal = {Cell Reports},
	number = {4},
	pages = {1050-1062.e6},
	title = {Translational Regulation of Non-autonomous Mitochondrial Stress Response Promotes Longevity},
	url = {https://www.sciencedirect.com/science/article/pii/S2211124719308587},
	volume = {28},
	year = {2019}}

@article{horenko_24b,
	article-number = {553},
	author = {Bassetti, Davide and Posp{\'\i}{\v s}il, Luk{\'a}{\v s} and Horenko, Illia},
	doi = {10.3390/e26070553},
	issn = {1099-4300},
	journal = {Entropy},
	number = {7},
	pubmedid = {39056915},
	title = {On Entropic Learning from Noisy Time Series in the Small Data Regime},
	url = {https://www.mdpi.com/1099-4300/26/7/553},
	volume = {26},
	year = {2024}}

@article{horenko_24a,
    author = {Vecchi, Edoardo and Bassetti, Davide and Graziato, Fabio and Pospíšil, Lukáš and Horenko, Illia},
    title = {Gauge-Optimal Approximate Learning for Small Data Classification},
    journal = {Neural Computation},
    volume = {36},
    number = {6},
    pages = {1198-1227},
    year = {2024},
    month = {05},
    issn = {0899-7667},
    doi = {10.1162/neco_a_01664},
    url = {https://doi.org/10.1162/neco\_a\_01664},
    eprint = {https://direct.mit.edu/neco/article-pdf/36/6/1198/2467115/neco\_a\_01664.pdf},
}

@article{TabPFN,
	author = {Hollmann, Noah and M{\"u}ller, Samuel and Purucker, Lennart and Krishnakumar, Arjun and K{\"o}rfer, Max and Hoo, Shi Bin and Schirrmeister, Robin Tibor and Hutter, Frank},
	date = {2025/01/01},
	doi = {10.1038/s41586-024-08328-6},
	id = {Hollmann2025},
	isbn = {1476-4687},
	journal = {Nature},
	number = {8045},
	pages = {319--326},
	title = {Accurate predictions on small data with a tabular foundation model},
	url = {https://doi.org/10.1038/s41586-024-08328-6},
	volume = {637},
	year = {2025}}

@article{ho2016generative,
  title={Generative adversarial imitation learning},
  author={Ho, Jonathan and Ermon, Stefano},
  journal={Advances in neural information processing systems},
  volume={29},
  year={2016}
}

@article{pfau2016connecting,
  title={Connecting Generative Adversarial Networks and Actor-Critic Methods},
  author={Pfau, Dirk and Viering, Koray and H{\"o}nlein, Ralf and Co-Reyes, Eric and Schmitt, Sebastian and Asfour, Tamim},
  journal={arXiv preprint arXiv:1611.07547},
  year={2016}
}

@inproceedings{ziebart2008maximum,
  title={Maximum entropy inverse reinforcement learning},
  author={Ziebart, Brian D and Maas, Andrew L and Bagnell, J Andrew and Dey, Anind K},
  booktitle={Proceedings of the 23rd AAAI conference on artificial intelligence},
  year={2008}
}

@inproceedings{haarnoja2018soft,
  title={Soft Actor-Critic: Off-Policy Maximum Entropy Deep Reinforcement Learning with a Stochastic Actor},
  author={Haarnoja, Tuomas and Zhou, Aurick and Abbeel, Pieter and Levine, Sergey},
  booktitle={International Conference on Machine Learning},
  year={2018}
}

@article{aggarwal21,
  title={Generative adversarial network: An overview of theory and applications},
  author={Aggarwal, Alankrita and Mittal, Mamta and Battineni, Gopi},
  journal={International Journal of Information Management Data Insights},
  volume={1},
  number={1},
  pages={100004},
  year={2021},
  publisher={Elsevier}
}

@inproceedings{li21b,
  title={Invisible backdoor attack with sample-specific triggers},
  author={Li, Yuezun and Li, Yiming and Wu, Baoyuan and Li, Longkang and He, Ran and Lyu, Siwei},
  booktitle={Proceedings of the IEEE/CVF international conference on computer vision},
  pages={16463--16472},
  year={2021}
}

@article{li21,
  title={Neural attention distillation: Erasing backdoor triggers from deep neural networks},
  author={Li, Yige and Lyu, Xixiang and Koren, Nodens and Lyu, Lingjuan and Li, Bo and Ma, Xingjun},
  journal={arXiv preprint arXiv:2101.05930},
  year={2021}
}

@inproceedings{wang23,
  title={Rfla: A stealthy reflected light adversarial attack in the physical world},
  author={Wang, Donghua and Yao, Wen and Jiang, Tingsong and Li, Chao and Chen, Xiaoqian},
  booktitle={Proceedings of the IEEE/CVF International Conference on Computer Vision},
  pages={4455--4465},
  year={2023}
}

@article{nguyen21,
  title={Imperceptible Warping-based Backdoor Attack. arXiv 2021},
  author={Nguyen, A and Tran, A WaNet},
  journal={arXiv preprint arXiv:2102.10369}
}

@article{manandhar14,
  title={Detection of faults and attacks including false data injection attack in smart grid using Kalman filter},
  author={Manandhar, Kebina and Cao, Xiaojun and Hu, Fei and Liu, Yao},
  journal={IEEE transactions on control of network systems},
  volume={1},
  number={4},
  pages={370--379},
  year={2014},
  publisher={IEEE}
}

@article{gu19,
  title={Badnets: Evaluating backdooring attacks on deep neural networks},
  author={Gu, Tianyu and Liu, Kang and Dolan-Gavitt, Brendan and Garg, Siddharth},
  journal={IEEE Access},
  volume={7},
  pages={47230--47244},
  year={2019},
  publisher={IEEE}
}

@inproceedings{bengio03,
  author = {Bengio, Yoshua and Paiement, Jean-François and Vincent, Pascal and Delalleau, Olivier and Roux, Nicolas Le and Ouimet, Marie},
  booktitle = {NIPS},
   editor = {Thrun, Sebastian and Saul, Lawrence K. and Schölkopf, Bernhard},
  ee = {http://papers.nips.cc/paper/2461-out-of-sample-extensions-for-lle-isomap-mds-eigenmaps-and-spectral-clustering},
  isbn = {0-262-20152-6},
  pages = {177-184},
  publisher = {MIT Press},
  title = {Out-of-Sample Extensions for LLE, Isomap, MDS, Eigenmaps, and Spectral Clustering. ObAfGkM12},
  url = {http://dblp.uni-trier.de/db/conf/nips/nips2003.html#BengioPVDRO03},
  year = 2003
}

@book{ilp_theory,
  author = {Schrijver, A.},
  url = {https://www.bibsonomy.org/bibtex/2c05d22504b2f52da8a72d854b5043004/wvdaalst},
  publisher = {Wiley-Interscience},
  title = {Theory of Linear and Integer programming},
  year = {1986},
}

@article{liu03,
  title={Audio watermarking techniques using sinusoidal patterns based on pseudorandom sequences},
  author={Liu, Zheng and Inoue, Akira},
  journal={IEEE Transactions on Circuits and Systems for Video Technology},
  volume={13},
  number={8},
  pages={801--812},
  year={2003},
  publisher={IEEE}
}

@article{swissroll_18,
	author = {Eldad Haber and Lars Ruthotto},
	doi = {10.1088/1361-6420/aa9a90},
	journal = {Inverse Problems},
	month = {dec},
	number = {1},
	pages = {014004},
	publisher = {IOP Publishing},
	title = {Stable architectures for deep neural networks},
	url = {https://dx.doi.org/10.1088/1361-6420/aa9a90},
	volume = {34},
	year = {2017}}

@InProceedings{adverse3,
  title = 	 {Adversarial Policies Beat Superhuman Go {AI}s},
  author =       {Wang, Tony Tong and Gleave, Adam and Tseng, Tom and Pelrine, Kellin and Belrose, Nora and Miller, Joseph and Dennis, Michael D and Duan, Yawen and Pogrebniak, Viktor and Levine, Sergey and Russell, Stuart},
  booktitle = 	 {Proceedings of the 40th International Conference on Machine Learning},
  pages = 	 {35655--35739},
  year = 	 {2023},
  editor = 	 {Krause, Andreas and Brunskill, Emma and Cho, Kyunghyun and Engelhardt, Barbara and Sabato, Sivan and Scarlett, Jonathan},
  volume = 	 {202},
  series = 	 {Proceedings of Machine Learning Research},
  publisher =    {PMLR},
  url = 	 {https://proceedings.mlr.press/v202/wang23g.html},
}

@article{adverse2,
	author = {Xu, Han and Ma, Yao and Liu, Hao-Chen and Deb, Debayan and Liu, Hui and Tang, Ji-Liang and Jain, Anil K.},
	date = {2020/04/01},
	doi = {10.1007/s11633-019-1211-x},
	id = {Xu2020},
	isbn = {1751-8520},
	journal = {International Journal of Automation and Computing},
	number = {2},
	pages = {151--178},
	title = {Adversarial Attacks and Defenses in Images, Graphs and Text: A Review},
	url = {https://doi.org/10.1007/s11633-019-1211-x},
	volume = {17},
	year = {2020}}

@unpublished{sparks23,
	author = {Bubeck, S{\'e}bastien and Chandrasekaran, Varun and Eldan, Ronen and Gehrke, Johannes and Horvitz, Eric and Kamar, Ece and Lee, Peter and Lee, Yin Tat and Li, Yuanzhi and Lundberg, Scott and Nori, Harsha and Palangi, Hamid and Ribeiro, Marco Tulio and Zhang, Yi},
	month = {March},
	title = {Sparks of Artificial General Intelligence: Early experiments with GPT-4},
	url = {https://www.microsoft.com/en-us/research/publication/sparks-of-artificial-general-intelligence-early-experiments-with-gpt-4/},
	year = {2023}}

@INPROCEEDINGS{adverse1,
  author={Chen, Jianbo and Jordan, Michael I. and Wainwright, Martin J.},
  booktitle={2020 IEEE Symposium on Security and Privacy (SP)}, 
  title={HopSkipJumpAttack: A Query-Efficient Decision-Based Attack}, 
  year={2020},
  volume={},
  number={},
  pages={1277-1294},
  doi={10.1109/SP40000.2020.00045}}

@article{horenko_pnas_23,
author = {Illia Horenko  and Edoardo Vecchi  and Juraj Kardoš  and Andreas Wächter  and Olaf Schenk  and Terence J. O’Kane  and Patrick Gagliardini  and Susanne Gerber },
title = {On cheap entropy-sparsified regression learning},
journal = {Proceedings of the National Academy of Sciences},
volume = {120},
number = {1},
pages = {e2214972120},
year = {2023},
doi = {10.1073/pnas.2214972120},
URL = {https://www.pnas.org/doi/abs/10.1073/pnas.2214972120},
eprint = {https://www.pnas.org/doi/pdf/10.1073/pnas.2214972120},
}

@article{Horenko_2020,
	doi = {10.1162/neco\_a\_01296},
	year = 2020,
	month = {aug},
	publisher = {{MIT} Press - Journals},
	volume = {32},
	number = {8},
	pages = {1563--1579},
	author = {Illia Horenko},
	title = {On a Scalable Entropic Breaching of the Overfitting Barrier for Small Data Problems in Machine Learning},
	journal = {Neural Computation}
}

@article{espa_22,
    author = {Vecchi, Edoardo and Pospíšil, Lukáš and Albrecht, Steffen and O'Kane, Terence J. and Horenko, Illia},
    title = "{eSPA+: Scalable Entropy-Optimal Machine Learning Classification for Small Data Problems}",
    journal = {Neural Computation},
    volume = {34},
    number = {5},
    pages = {1220-1255},
    year = {2022},
    month = {04},
    issn = {0899-7667},
    doi = {10.1162/neco_a_01490},
    url = {https://doi.org/10.1162/neco\_a\_01490},
    eprint = {https://direct.mit.edu/neco/article-pdf/34/5/1220/2008663/neco\_a\_01490.pdf},
}

@article{horenko_pnas_22,
author = {Illia Horenko },
title = {Cheap robust learning of data anomalies with analytically solvable entropic outlier sparsification},
journal = {Proceedings of the National Academy of Sciences},
volume = {119},
number = {9},
pages = {e2119659119},
year = {2022},
doi = {10.1073/pnas.2119659119},
URL = {https://www.pnas.org/doi/abs/10.1073/pnas.2119659119},
eprint = {https://www.pnas.org/doi/pdf/10.1073/pnas.2119659119},
}

@article{Gerber_2020,
	doi = {10.1126/sciadv.aaw0961},
	year = 2020,
	month = {jan},
	publisher = {American Association for the Advancement of Science ({AAAS})},
	volume = {6},
	number = {5},
	pages = {eaaw0961},
	author = {S. Gerber and L. Pospisil and M. Navandar and I. Horenko},
	title = {Low-cost scalable discretization, prediction, and feature selection for complex systems},
	journal = {Science Advances}
}

@article{horenko17,
	author = {Susanne Gerber and Illia Horenko},
	doi = {10.1073/pnas.1612619114},
	eprint = {https://www.pnas.org/doi/pdf/10.1073/pnas.1612619114},
	journal = {Proceedings of the National Academy of Sciences},
	number = {19},
	pages = {4863-4868},
	title = {Toward a direct and scalable identification of reduced models for categorical processes},
	url = {https://www.pnas.org/doi/abs/10.1073/pnas.1612619114},
	volume = {114},
	year = {2017}}

\end{document}